%% file: main.tex
\newcommand{\ignore}[1]{}

\documentclass[conference]{IEEEtran}
\IEEEoverridecommandlockouts
\usepackage{cite}
\usepackage{amsmath,amssymb,amsfonts}
\usepackage{algorithmic}
\usepackage{graphicx}
\usepackage{textcomp}
\usepackage{xcolor}
\usepackage{caption}
\usepackage{subcaption}
\def\BibTeX{{\rm B\kern-.05em{\sc i\kern-.025em b}\kern-.08em
    T\kern-.1667em\lower.7ex\hbox{E}\kern-.125emX}}
\begin{document}

\title{Energy-Efficient High-Accuracy\\Spiking Neural Network Inference\\Using Time-Domain Neurons
\thanks{This research was supported in part by National R\&D Program through the National Research Foundation of Korea (NRF) funded by Ministry of Science and ICT (2021M3F3A2A01037928) and by Ministry of Science and ICT (2020M3H2A1078119).}
}
\setlength{\skip\footins}{1em}

\author{
    %\IEEEauthorblockN{
        Joonghyun Song\IEEEauthorrefmark{1}, Jiwon Shin\IEEEauthorrefmark{1}, Hanseok Kim\IEEEauthorrefmark{1}\IEEEauthorrefmark{2}, and Woo-Seok Choi\IEEEauthorrefmark{1}\\
    %}
    %\IEEEauthorblockA{    
    %    \textit{
            \IEEEauthorrefmark{1}\textit{Department of ECE, ISRC, Seoul National University,} Seoul, South Korea \\
            \IEEEauthorrefmark{2}\textit{Samsung Electronics,} Hwaseong, South Korea  \\ 
            Email: \{jhsong1997, wooseokchoi\}@snu.ac.kr\vspace{-1em}
    %    }
    %} 
}

\ignore{
\author{\IEEEauthorblockN{Joonghyun Song}
\IEEEauthorblockA{\textit{Dept of ECE, ISRC} \\
\textit{Seoul National University}\\
Seoul, South Korea \\
jhsong1997@snu.ac.kr\vspace{-2em}}
\and
\IEEEauthorblockN{Jiwon Shin}
\IEEEauthorblockA{\textit{Dept of ECE, ISRC} \\
\textit{Seoul National University}\\
Seoul, South Korea \\
sjw7323@snu.ac.kr\vspace{-2em}}
\and
\IEEEauthorblockN{Hanseok Kim}
\IEEEauthorblockA{\textit{Dept of ECE, ISRC} \\
\textit{Seoul National University}\\
\textit{Samsung Electronics}\\
Seoul, South Korea \\
anjeo@snu.ac.kr\vspace{-2em}}
\and
\IEEEauthorblockN{Woo-Seok Choi}
\IEEEauthorblockA{\textit{Dept of ECE, ISRC} \\
\textit{Seoul National University}\\
Seoul, South Korea \\
wooseokchoi@snu.ac.kr\vspace{-2em}}
}
}

\maketitle

\input{Abstract}
\input{0_Introduction}
\input{1_Time}
\input{2_Simulation}
\input{3_Conclusion}

\section*{Acknowledgment}
The EDA Tool was supported by the IC Design Education Center.

\bibliographystyle{IEEEtran}
\bibliography{Ref}

\end{document}

%% file: Abstract.tex
\begin{abstract}
Due to the limitations of realizing artificial neural networks on prevalent von Neumann architectures, 
recent studies have presented neuromorphic systems based on spiking neural networks (SNNs) to reduce power and computational cost.
However, conventional analog voltage-domain integrate-and-fire (I\&F) neuron circuits, based on either current mirrors or op-amps, 
pose serious issues such as nonlinearity or high power consumption, thereby degrading either inference accuracy or energy efficiency of the SNN.
To achieve excellent energy efficiency and high accuracy simultaneously, this paper presents a low-power highly linear time-domain I\&F neuron circuit.
Designed and simulated in a 28\,nm CMOS process, the proposed neuron leads to more than 4.3\,$\times$ lower error rate on the MNIST inference over the conventional current-mirror-based neurons.
In addition, the power consumed by the proposed neuron circuit is simulated to be 0.230\,$\mu$W per neuron, which is orders of magnitude lower than the existing voltage-domain neurons.
\end{abstract}

\begin{IEEEkeywords}
artificial neural network, spiking neural network, ANN-to-SNN conversion, integrate-and-fire neuron, time-domain signal processing
\end{IEEEkeywords}
\vspace{-1em}

%% file: 0_Introduction.tex
\section{Introduction}
\label{sec:intro}

Artificial neural networks (ANNs) have been utilized to achieve groundbreaking results in a variety of fields such as image recognition~\cite{lecun1998gradient}, speech recognition~\cite{hinton2012deep}, and machine translation~\cite{wu2016google}.
However, implementing such networks with a large number of parameters on conventional von Neumann architectures 
incurs tremendous latency and energy costs dominated by memory access, which limits the use of deep networks in mobile applications~\cite{horowitz20141}.
Hence, recent studies have focused on developing a new type of system that can replace the prevalent architectures.
Especially, neuromorphic systems using spiking neural networks (SNNs) are considered an alternative since they are effective in reducing both power consumption and computational effort~\cite{seo201145nm}.
While there have been a multitude of researches on direct SNN training methods, 
SNNs based on simple integrate-and-fire (I\&F) neurons with the ANN-to-SNN conversion technique~\cite{rueckauer2017conversion} show better inference accuracy compared to other models.
Thus, this paper focuses on the design of a novel I\&F neuron that leads to high-accuracy SNN inference in an energy-efficient manner.
%A hardware implementation of SNN can be done by using integrate-and-fire (I\&F) neurons.
%I\&F neurons receive pre-synaptic spikes from the preceding layer and accumulate them as the membrane potential.
%When the membrane potential exceeds a threshold, a spike is generated by the neuron and delivered to the next layer.
%In addition, SNNs based on I\&F neurons with the ANN-to-SNN conversion technique~\cite{rueckauer2017conversion} show better inference accuracy compared to other models.
%Thus, this paper focuses on the design of a novel I\&F neuron that leads to high-accuracy SNN inference in an energy-efficient manner.

Since conventional voltage-domain analog neurons exhibit nonlinear behavior that causes inference accuracy degradation, 
\cite{kim2022improving} proposes a time-domain neuron, which improves linearity, thereby achieving inference accuracy on a par with that of ANNs.
However, questions on how to design a low-power time-domain neuron and, more importantly, whether they can be used to implement an SNN achieving good energy efficiency and high accuracy simultaneously,  remain unclear.
To answer these questions, this paper presents a low-power design of the time-domain I\&F neuron and simulates an SNN for MNIST inference using the proposed neuron.
Implemented in a 28\,nm CMOS process, the performance of the proposed neuron is far superior to that of existing analog neurons while operating energy-efficiently.

%% file: 1_Time.tex
\section{Time-Domain Neuron}
\label{sec:time}

\captionsetup{labelsep=period}
\begin{figure*}
    \centering
    \includegraphics[scale=0.24]{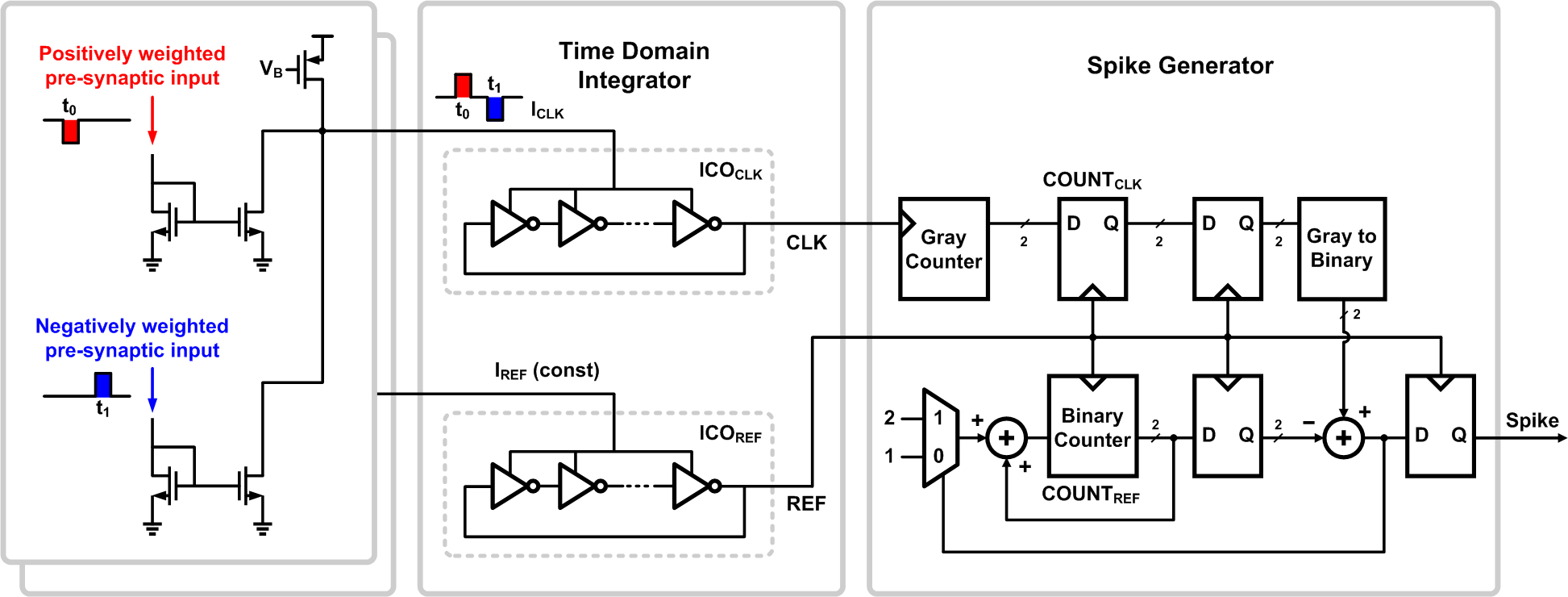}
    \caption{Proposed architecture of the low-power mixed-signal time-domain I\&F neuron.}
    \label{fig:block}
    \vspace{-1em}
\end{figure*}

\subsection{Limitations of Voltage-Domain Neurons}
Current-mirror-based neurons~\cite{indiveri2006vlsi,hwang2020impact} and op-amp-based neurons~\cite{aamir2018accelerated,asghar2021current} are the two types of analog I\&F neurons commonly used in hardware-based SNNs.
The I\&F neurons receive pre-synaptic spikes from the preceding layer and accumulate them as membrane potential in the voltage domain.
When the membrane potential exceeds a threshold, a spike is generated by the neuron and delivered to the next layer.
These voltage-domain neurons suffer from serious issues in either performance or power.
Since the membrane potential, or the voltage of the capacitor, changes the amount of current from the current mirror that flows into the capacitor due to the channel length modulation, nonlinear behavior is observed in the current-mirror-based neuron~\cite{kim2022improving}.
This leads to the performance degradation of the SNN, which is later validated in Section~\ref{sec:simulation}.
On the other hand, op-amp-based neurons require high supply voltage to provide sufficient gain and linearity, 
thereby suffering from high power consumption.

The design challenges of the voltage-domain neurons get worse as technology scales down and the supply voltage reduces.
Due to the lower output impedance/intrinsic gain and higher leakage of the devices, 
nonlinear behavior in the voltage-domain neurons becomes severe in deep sub-micron technologies.
Moreover, as lower supply voltage reduces both voltage headroom and the range of membrane potential, performance degradation of the voltage-domain neurons is inevitable.
These issues can be partly overcome by choosing large devices to design neurons, 
but this incurs large area/power overhead, limiting the usage of deeper and larger networks.

\subsection{Time-Domain Signal Processing}
Time-domain signal processing has emerged as an attractive technique to mitigate the analog circuit design challenges associated with process scaling and reduced voltage headroom, 
and it has been applied to analog filters~\cite{drost2012analog}, data converters~\cite{reddy201216}, and PID controllers~\cite{kim201710}.
The main idea behind time-domain signal processing is to encode the variable of interest in the time, or phase, domain, which fundamentally solves the voltage-headroom-related issues.
For instance, since phase is obtained by integrating frequency with respect to time,
an ideal integrator of a voltage signal can be implemented in the phase domain using a voltage controlled oscillator (VCO) 
whose frequency is controlled by the voltage signal.
An I\&F neuron exploiting this concept is presented in \cite{kim2022improving}, which shows highly linear behavior and better performance compared to the existing current-mirror-based voltage-domain neurons.

In the time-domain neuron, whenever a spike is received from the preceding layer, a current pulse flows into the current mirror.
Depending on the polarity of the synaptic weights, 
the amount of current that flows into the current controlled oscillator (ICO) either increases or decreases.
This changes the ICO frequency, and a phase is shifted accordingly.
Since the amount of phase shift is the accumulation of the weight-multiplied input spikes, 
the time-domain neuron generates and transmits a spike to the following layer when the phase shift reaches a threshold.
Although time-domain neurons can potentially solve many design issues existing in voltage-domain neurons, 
an important question on whether both excellent energy efficiency and high inference accuracy can be achieved simultaneously using time-domain neurons has not been answered yet.

\subsection{Low-Power Design of Time-Domain Neuron}
To answer the question, 
we first propose an ultra-low-power mixed-signal time-domain I\&F neuron circuit (see Fig.~\ref{fig:block}).
When a spike is received from the preceding layer, a current pulse (typically generated by synaptic devices like memristors~\cite{chu2014neuromorphic}) 
is supplied from the synapse array and flows into the neuron circuit.
Note that when an input spike is modulated with a positive (negative) weight, the ICO frequency should increase (decrease).
A straightforward way to implement this is to use a current mirror as \cite{hwang2020impact}, where PMOS/NMOS are used to add/subtract current flowing into the ICO, respectively.
However, this implementation suffers from a mismatch between PMOS and NMOS, which manifests itself as weight error and leads to inference accuracy degradation.
To circumvent this issue, the proposed neuron uses only NMOSs to increase/decrease the ICO current as shown in Fig.~\ref{fig:block}.
For positive weights, less current flows into the current mirror\footnote{This can be implemented by having current from the synapse constantly flow (steady state) and then gating the current when receiving an input spike.}, 
while more current flows in for negative weights.
This results in an increase or decrease, respectively, in the ICO current (i.e. ICO frequency) as the constant current is provided by the PMOS current source.
With this time-domain neuron, the membrane potential is now embedded in the ICO phase.
%\ignore{
%\begin{figure}
%    \centering
%    \includegraphics[width=\columnwidth]{resetblock.png}
%    \caption{Block diagram of the proposed digital spike generator.}
%    \label{fig:reset block}
%    \vspace{-1em}
%\end{figure}
%}

To detect the amount of phase shift, which represents the membrane potential, a reference phase is required.
The proposed neuron uses an identical ICO whose current, or frequency, is fixed to generate a reference.
Note that, since this reference ICO can be shared for all the neurons, it adds negligible area overhead.
Two counters are used to detect whether the phase difference between the two ICO outputs, $CLK$ and $REF$, crosses a threshold by counting the numbers of rising edges.
In our implementation, when the phase difference $\Phi_{CLK} - \Phi_{REF}$ crosses the threshold $2\pi$, the difference between the counter outputs becomes 1.
Then, a digital spike generator (see Fig.~\ref{fig:block}) generates a spike and transmits it to the following layer.
Since two counters are synchronized with different clocks, 
gray code with cascaded flip-flops is used for clock domain crossing.

\begin{figure}
    \centering
    \includegraphics[width=\columnwidth]{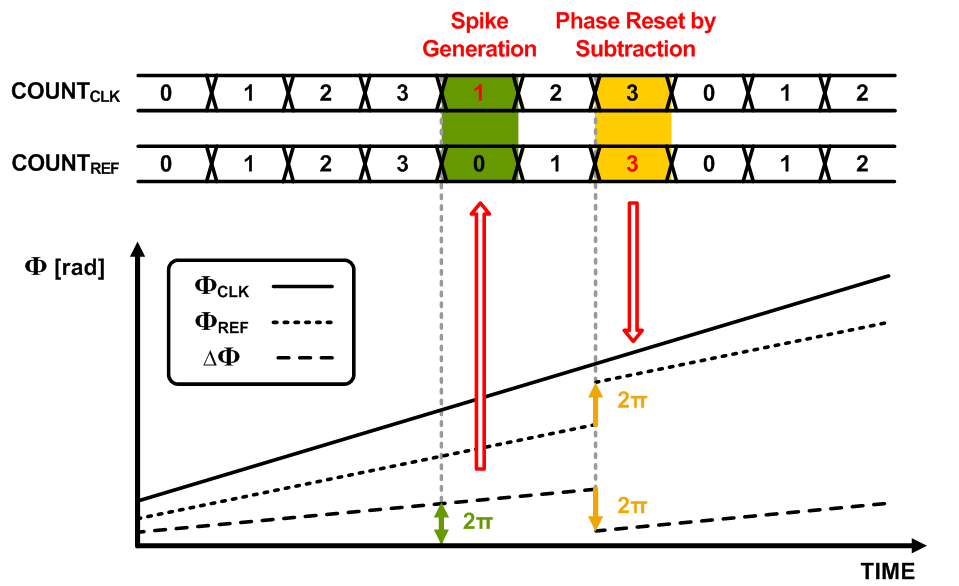}
    \caption{Spike generation and reset by subtraction mechanism.}
    \label{fig:reset waveform}
    \vspace{-1em}
\end{figure}

The proposed time-domain I\&F neuron circuit does not need any extra hardware to implement the \textit{``reset by subtraction''} mechanism, 
which enables more accurate inference than ``reset to zero'' when converting ANNs into SNNs~\cite{rueckauer2017conversion}.
When a spike is generated, 1 is added to the reference counter value to reset the phase difference, as shown in Fig.~\ref{fig:reset waveform}, 
by which the exact amount of the threshold ($2\pi$ phase) can be subtracted from the original phase difference.
In the proposed design, in order to minimize the number of bits in digital circuits, 
which incurs minimal area/power overhead, 
the threshold is set to $2\pi$ radians in phase, 
equivalent to the counter output difference of 1.
%because the number of bits in digital circuits can be minimized, 
%incurring minimal area/power overhead.
Moreover, since the proposed neuron is highly digital, 
energy efficiency can be improved substantially by lowering the supply voltage down to a range of 0.35\,V-0.5\,V, 
which is inapplicable to the existing analog voltage-domain neurons.

%% file: 2_Simulation.tex
\section{Simulation Results}
\label{sec:simulation}

\subsection{Operation of Proposed Time-Domain I\&F Neuron}
\begin{figure} [t]
    \centering
    \includegraphics[width=\columnwidth]{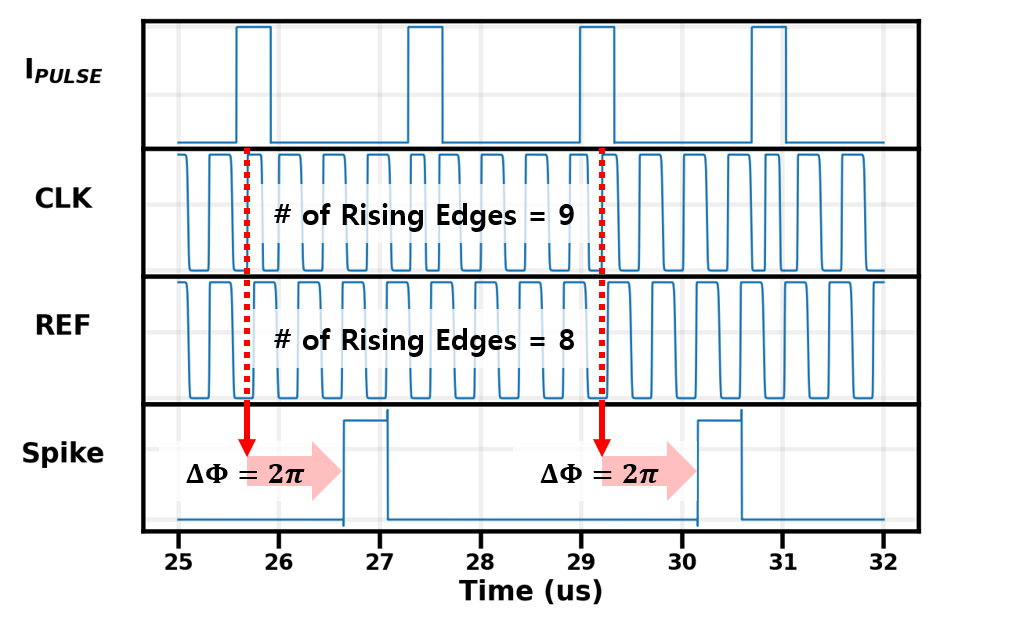}
    \caption{Simulated waveform of the proposed neuron.}
    \label{fig:waveform}
    \vspace{-1em}
\end{figure}

The proposed time-domain I\&F neuron circuit is designed and simulated in a 28\,nm CMOS process.
The simulated waveforms of the neuron are plotted in Fig.~\ref{fig:waveform}.
When the current pulses are periodically supplied from the synapse array, the phase difference between two ICOs, $\Delta\Phi = \Phi_{CLK} - \Phi_{REF}$, eventually becomes larger than $2\pi$, 
which can be detected by comparing the numbers of rising edges of $CLK$ and $REF$, counted by two counters.
If the numbers differ as shown in Fig.~\ref{fig:waveform}, a spike is generated and the phase difference is subtracted by $2\pi$ by resetting the counters.
From the supply voltage ranging from 0.35\,V to 0.5\,V, the designed ICO oscillates with the frequency from 3\,MHz to 100\,MHz, which is also the range of the maximum firing rates of the neuron.

\begin{figure} [t]
    \centering
    \includegraphics[width=\columnwidth]{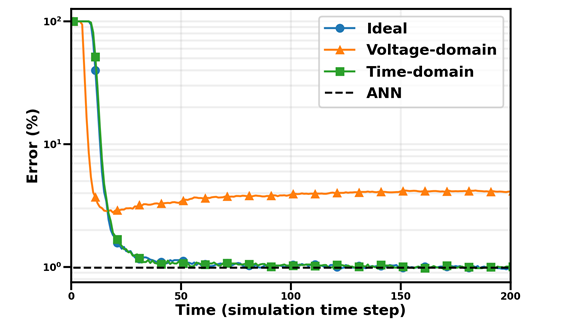}
    \caption{Simulated inference error rate on MNIST for different neurons.}
    \label{fig:error-simtimestep}
    \vspace{-1em}
\end{figure}
\begin{figure} [hbt!]
    \centering
    \includegraphics[width=\columnwidth]{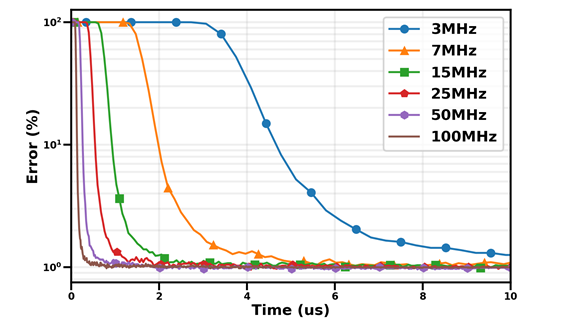}
    \caption{Simulated inference error rate versus latency for the proposed neurons with different maximum firing rates.}
    \label{fig:error-time}
    \vspace{-1em}
\end{figure}

\subsection{SNN System Simulation with Proposed Neuron}
To evaluate the performance of the SNN using the proposed time-domain neurons, LeNet-5~\cite{lecun1998gradient} is trained on the MNIST dataset and converted to the SNN by extracting the network parameters using the ANN-to-SNN conversion tool~\cite{rueckauer2017conversion}.
Fig.~\ref{fig:error-simtimestep} displays the inference error rate on MNIST with respect to the simulation time step for three different neurons: 
the ideal I\&F neuron, the current-mirror-based voltage-domain neuron, and the proposed time-domain neuron.
The simulated error rates of these three neurons are 0.99\,\% (ideal), 4.31\,\% (voltage-domain), and 0.98\,\% (proposed), respectively.
While the accuracy of the time-domain neuron approaches that of the ideal I\&F neuron and the original ANN (0.99\,\%), 
the voltage-domain neuron fails to reach the same accuracy level due to its nonlinearity.
In Fig.~\ref{fig:error-time}, the error rates for the time-domain neurons with 6 different oscillation frequencies, or maximum firing rates (from 3\,MHz to 100\,MHz), are illustrated, 
where the x-axis represents the absolute latency in microseconds.
The result indicates that, although all the neurons eventually approach the same accuracy level, lower inference latency can be achieved when the maximum firing rate becomes higher.

\begin{figure} [t]
    \centering
    \includegraphics[width=\columnwidth]{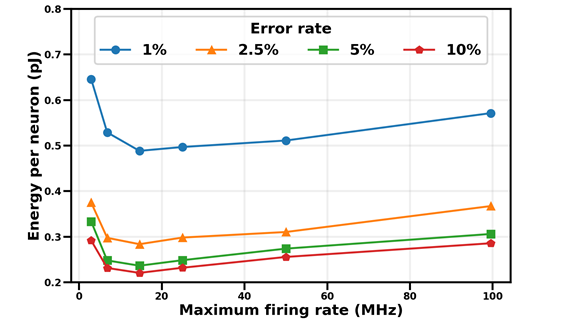}
    \caption{Average energy per inference that each neuron with different maximum firing rate consumes to reach the error rates of 1\,\%, 2.5\,\%, 5\,\%, and 10\,\%.}
    \label{fig:energy}
\end{figure}

\setlength{\tabcolsep}{3pt}
\renewcommand{\arraystretch}{1.25}
\begin{table}%[t]
\centering
\caption{Performance summary and comparison.}
\begin{tabular}{| c | c | c | c | c |} 
 \hline
  & \cite{indiveri2006vlsi} & \cite{aamir2018accelerated} & \cite{asghar2021current} & \textbf{This work} \\ [0.5ex] 
 \hline\hline
 Process & 800\,nm & 65\,nm & 65\,nm & \textbf{28\,nm} \\ 
 Domain & Voltage & Voltage & Voltage & \textbf{Time} \\ 
 Key element & Current mirror & Op-amp & Op-amp & \textbf{ICO} \\
 Power per neuron & $>$~10\,$\mu$W  & 14.4\,$\mu$W  & 3\,$\mu$W & \textbf{0.230\,$\mu$W} \\
 Energy per inference & N/A & N/A & N/A & \textbf{3.72\,nJ} \\
 Inference accuracy & N/A & N/A & N/A & \textbf{99\,\%} \\ [1ex]
 \hline
\end{tabular}
\label{table:1}
\vspace{-1em}
\end{table}

\subsection{Energy Efficiency of Proposed Neuron}
Fig.~\ref{fig:energy} shows the total energy consumption per neuron for each inference to achieve a certain level of accuracy for the proposed neuron with different maximum firing rates.
Neurons with lower ICO frequency result in reduced power consumption compared to those with higher frequency.
However, as shown in Fig.~\ref{fig:energy}, since more time should be spent to achieve the same level of accuracy (see Fig.~\ref{fig:error-time}), 
less power does not simply imply lower energy for inference.
For example, the lowest total energy per neuron to achieve 1\,\% error is 0.488\,pJ at 15\,MHz ICO frequency while the highest is 0.646\,pJ at 3\,MHz.
It is important to notice that there is an optimal operating point that minimizes the total energy consumption, e.g. 15\,MHz operating frequency in our simulation results.
Therefore, when choosing the operating frequency of the time-domain neuron, both total energy consumption and latency should be carefully considered.

Table~\ref{table:1} summarizes the performance of the proposed neuron and compares it with the prior art.
The proposed neuron shows the power consumption of 0.230\,$\mu$W, which is substantially low compared to the voltage-domain neurons.
Note that, as technology scales down, the voltage-domain neurons will suffer from either higher power consumption or degraded inference accuracy due to nonlinearity, 
and the benefits of the proposed time-domain neuron are expected to become larger.
The SNN using the proposed neuron consumes only 3.72\,nJ for classifying an MNIST image while achieving 99\,\% inference accuracy.

%% file: 3_Conclusion.tex
\section{Conclusion}
\label{sec:conclusion}

This paper proposes a low-power highly linear I\&F neuron circuit composed of a time-domain integrator and a digital spike generator.
Designed and simulated in a 28\,nm CMOS process, the proposed neuron leads to more than $4.3\,\times$ lower error rate on the MNIST inference over the conventional analog current-mirror-based neurons.
In addition, the power consumed by the proposed neuron circuit is 0.230\,$\mu$W per neuron, which is orders of magnitude lower than the existing voltage-domain neurons.
The simulation results indicate that the proposed time-domain neuron enables the SNN inference to achieve excellent energy efficiency and high accuracy.% simultaneously.